\def\year{2020}\relax
\documentclass[letterpaper]{article} 
\usepackage{aaai20}  
\usepackage{times}  
\usepackage{helvet} 
\usepackage{courier}  
\usepackage[hyphens]{url}  
\usepackage{graphicx} 
\usepackage{graphicx}  
\usepackage{comment}
\usepackage[boxed,ruled,commentsnumbered]{algorithm2e}
\usepackage{amssymb,amsmath,amsthm}

\usepackage{booktabs}
\usepackage{pifont}
\usepackage[caption=false]{subfig}
\newcommand{\citet}[1]{\citeauthor{#1} \shortcite{#1}}
\usepackage{bbm}
\usepackage{array}
\usepackage{multirow}
\newcounter{tightlistcnt}
\newenvironment{tightitemize}{%
	\begin{list}{\textbullet}{\usecounter{tightlistcnt}%
			\topsep 0in
			\partopsep 0in
			\itemsep .2em
			\parsep 0in
			\leftmargin 0em
			\rightmargin 0in
			\listparindent 0em
			\itemindent .75em
			\labelsep .75em
			\labelwidth 0in
		}%
	}{%
	\end{list}%
}

\newenvironment{tightenumerate}{%
	\begin{list}{\arabic{tightlistcnt}.}{\usecounter{tightlistcnt}
			\topsep 0in
			\partopsep 0in
			\itemsep .2em
			\parsep 0in
			\leftmargin 0.0em
			\rightmargin 0in
			\listparindent 0em
			\itemindent .75em
			\labelsep .75em
			\labelwidth 0in
		}%
	}{%
	\end{list}%
}

\title{Unlocking the Power of Deep PICO Extraction: Step-wise Medical NER Identification}

\author{Tengteng Zhang$^{1}$, Yiqin Yu$^{2}$, Jing Mei$^{2}$, Zefang Tang$^{2}$, Xiang Zhang$^{1}$, Shaochun Li$^{2}$\\
	\\
	$^1$Southeast University, Nanjing, China;
	$^2$IBM Research, Beijing, China\\ 
}
\begin{document}
	
	\maketitle
	
	\begin{abstract}
		
		The PICO framework (Population, Intervention, Comparison, and Outcome) is usually used to formulate evidence in the medical domain. The major task of PICO extraction is to extract sentences from medical literature and classify them into each class. However, in most circumstances, there will be more than one evidences in an extracted sentence even it has been categorized to a certain class. In order to address this problem, we propose a step-wise disease Named Entity Recognition (DNER) extraction and PICO identification method. With our method, sentences in paper title and abstract are first classified into different classes of PICO, and medical entities are then identified and classified into P and O. Different kinds of deep learning frameworks are used and experimental results show that our method will achieve high performance and fine-grained extraction results comparing with conventional PICO extraction works.
		
	\end{abstract}
	
	\section{Introduction}
	\indent\setlength{\parindent}{1em}Evidence-based medicine (EBM) is a dominated approach enabling clinical practitioners to utilize the best available information (evidence) in making decision during clinical practice. When seeking evidences in medical literature, the PICO framework \cite{schardt2007utilization} is usually used to formulate evidence in the medical domain. PICO represents four elements: Patient/Problem (P), Intervention (I), Comparison (C) and Outcome (O). Unfortunately, the PICO elements are usually not clearly identified in the text contents of medical literature, which need clinical practitioners to read and extract the PICO manually \cite{yuan_extracting_2019}. 
	
	The major task of automatic PICO extraction is to extract sentences from medical literature and classify them into each element (class).
	Most of the previous methods \cite{jin_pico_2018} identify PICO elements in abstracts and do classification at the sentence level.
	Unfortunately in most circumstances, there will be more than one PICO elements in an extracted sentence even it has been categorized to one class with the highest probability. So the fine-grained entity level extraction for PICO elements other than sentence level PICO elements is required. 
	
	In this paper, we propose a fine-grained medical Named Entity Recognition (NER) extraction and PICO identification method. With our method, medical NERs are identified and categorized into different classes of PICO.
	There are a lot of works using natural language processing (NLP) algorithms to extract medical NERs and classify them into different classes. Many deep neural network architectures have been proposed and used, among which the Conditional random field layer combined with the model of bidirectional Long Short-Term Memory (BiLSTM-CRF) produces encouraging results on medical NER extraction \cite{huang_bidirectional_2015}. Conditional Random Field (CRF) is a graph model that is often used to solve sequence labeling problems which focus on sentence level positions. Compared to the softmax classifier, the CRF can capture the context information of the target label. Further, the Convolutional Neural Networks (CNNs) and LSTM networks are applied to encode character-level word embedding. In addition, there are many public knowledge bases in the medical field which are used to enhance the deep learning frameworks. Previous works have compared the performance of these models. However, there are still a lack of validation on classifying the extracted medical NERs into PICO classes.
	
	This paper tries to leverage the dominated deep learning technologies such as Bi-LSTM, CNN and BERT \cite{devlin_bert:_2018,alsentzer_publicly_2019} and properly position them into different steps to streamline the overall entity level PICO identification task. Structured information in medical knowledge graph such as UMLS are also used to enhance the performance. The main contribution of this paper are as follows: 
	\begin{itemize}
		\item We propose a step-wise work flow to fulfill the disease entity classification task, which carefully position different deep learning models into different steps. It makes us to improve the performance step-by-step and enhance the explainability of the whole system.
		\item We introduce mainstream deep learning models for the sentence-level PICO classification task and the disease entity recognition task. Structured information of medical domain knowledge via graph embedding is integrated into the deep learning models. Language rules are also combined with deep methods to provide adjusted entity-level classification results. 
		\item A series of experiments for each step are implemented to show step-wise results. A web application is developed to enable the interaction between these results and the end users. Results on a set of risk model papers are also discussed.
	\end{itemize}
	
	The method is evaluated on clinical predictive modeling papers \cite{steyerberg2009clinical}, which is a major category of medical literature. In these papers, risk models are proposed as an explicit, empirical approach to estimate probabilities of disease(s) or outcome(s). As a powerful tool of disease prevention, risk models are widely used in chronic disease management. Annually, about two-thirds of total deaths around the world are caused by chronic diseases \cite{oh_development_2014}. Half of them are cardiovascular diseases. Risk models are essential for physicians to estimate individual patient’s risk or discover leading risk factors. However, there are several limitations among risk models, such as low sensitivity, inadequate factors in model\cite{elosua_cardiovascular_2014} or not robust across domains. To automatically extract medical entities and classify them into PICO will be the first step to evaluate and utilize these risk models. To be specific, we address the disease NER extraction and classification in this paper.
	
	\section*{Method}
\begin{figure*}[h!]
		\centering
		\includegraphics[scale=1, width=420pt]{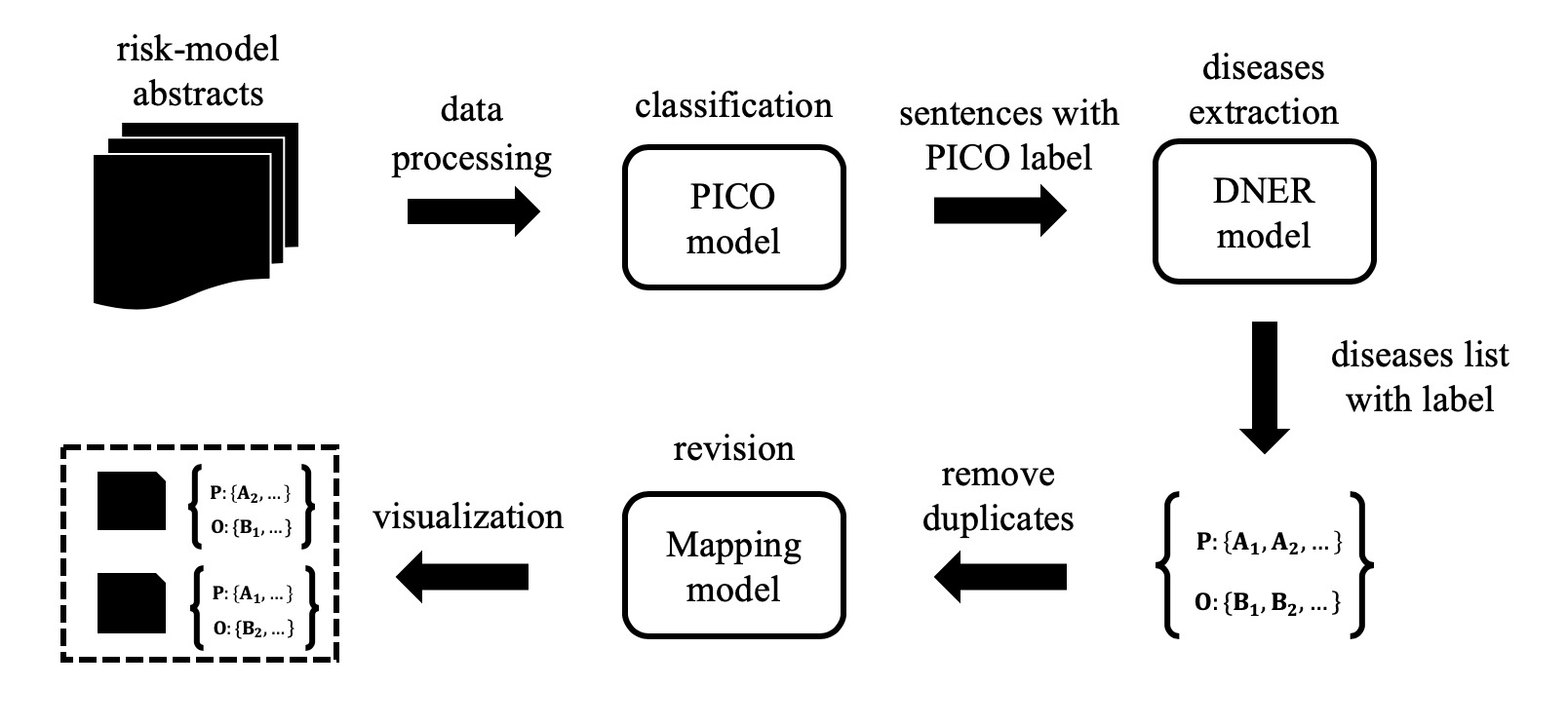}
		\caption{The system work flow. There are four main modules: PICO classification model, diseases named entity recognition model, mapping model and visualization system.}
		\label{fig1}
	\end{figure*}

	\noindent The popular end-to-end approach for PICO entity extraction is mainly in black-box which reduces the explainability of the model. To address this issue, we divide the complex system into different components. The main work flow of our method is illustrated in Figure ~\ref{fig1}.  Text content of title and abstract of risk model papers are used as inputs, where the papers are retrieved by our previous work \cite{yu_method_nodate}. In a nutshell, we provide a step-wise work flow to extract and classify disease entities in risk model paper title and abstract into PICO classes. Sentences in title and abstract are firstly classified into PICO classes by the PICO Classification Model. Following with the DNER Model, disease entities are then extracted from the sentences. Finally, the Disease Entity Mapping Model adjust and re-classify the disease entities into P and O. There are several advantages for this framework. First, the entity classification task is divided into different tasks, e.g., sentence-level and entity-level, which makes it possible to leverage different kinds of deep learning models and improve the performance step-by-step. Second, since human interaction can not be avoided, sophistically dividing the task into several steps with meaningful inputs and outputs will improve the explainability of the whole system. In our paper, a web application is also shown to demonstrate how the results of each step are reported to the end user, and how these results can be modified via the user interface.  
	
	\subsection*{PICO Classification Model}
	
\begin{figure*}[h!]
		\centering
		\includegraphics[scale=1, width=420pt]{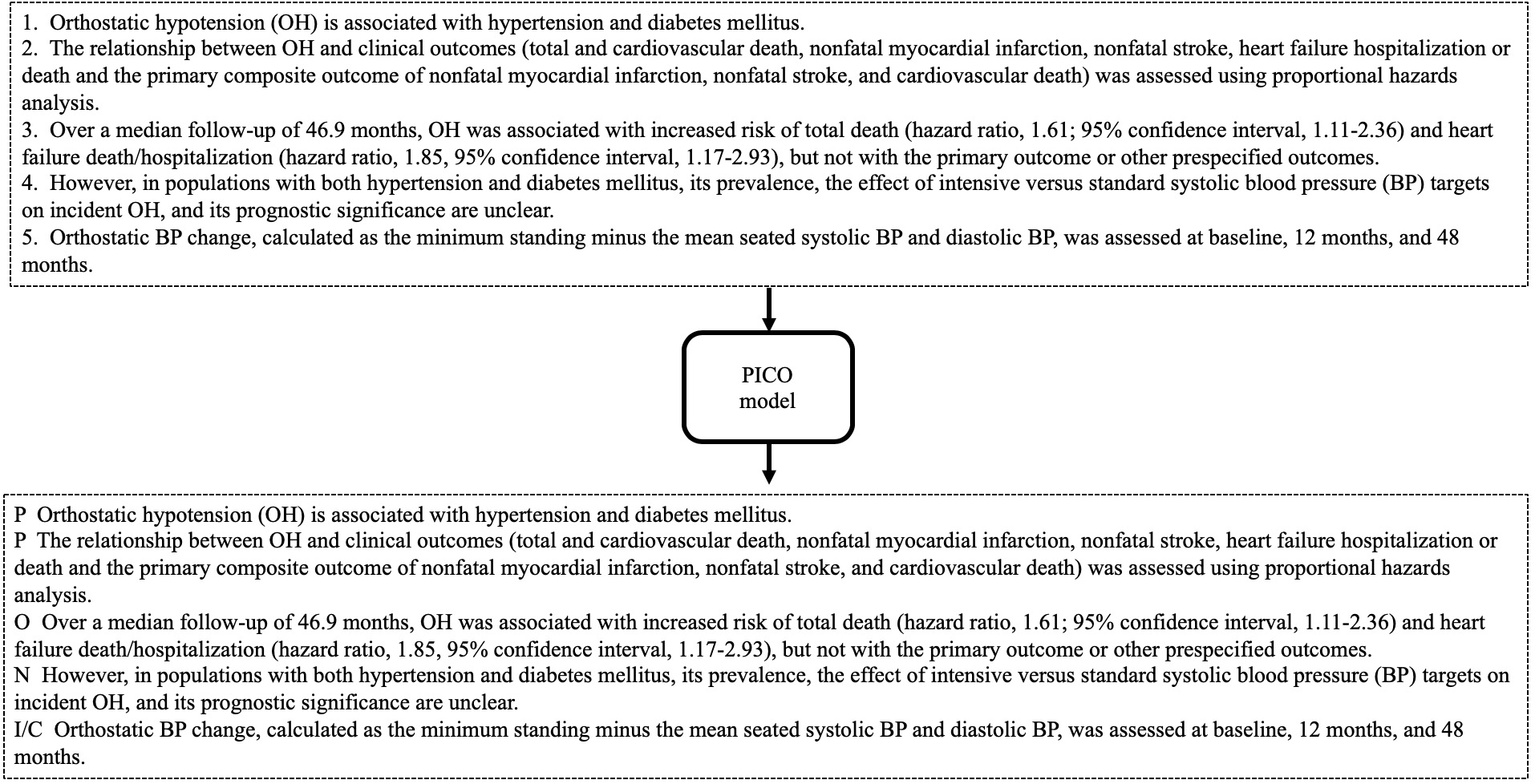}
		\caption{The PICO sentences classification model. Its input is the sentences of paper and outputs are sentences with labels.}
		\label{fig2}
	\end{figure*}

	\noindent The PICO classification model is generally a sentence classification model. Its input is the title and abstract of each paper. 
	The output is sentences along with the PICO classes including P, I/C, O, and N. Label N means that this sentence doesn't belong to any class of PICO.
	Like many other works \cite{wallace_extracting_nodate,jin_pico_2018}, we incorporate the element C into the element I since the ``comparison'' usually refers to a kind of ``intervention'' in most literature. 
	The sentence classification example is shown in Figure \ref{fig2}. 
	
	Traditional sentence classification models are often based on machine learning techniques, which need complex feature engineering and have poor transferability\cite{pan_survey_2010}.
	The popular deep learning models have changed this situation, and excellent models combined with large high-quality data have generated amazing results \cite{kim_convolutional_2014}.
	One of the major shortcomings of other paper works is the lack of high quality annotated dataset about PICO sentences. The dataset is either not public or coarse-grained labelled. In order to solve this problem, we invite experts to label 500 papers which will be described in experiment part. Two classification models are implemented: CNN (Convolutional Neural Network)  \cite{kim_convolutional_2014,yang_hierarchical_2016}and Bi-LSTM (Bidirectional Long Short-Term Memory)\cite{graves_framewise_2005}.
	The most classic model for text classification by CNN is the work of kim\cite{kim_convolutional_2014}. 
	Its input layer is a word vector expression obtained by passing the public corpus through the pre-trained word2vec\cite{mikolov_distributed_nodate} model, and the output layer is the sentence classification label probability. The input word sequence can be transformed into a vector by concatenating the corresponding word vectors from the embedding matrix. 
	The word embedding we use is trained under large-scale medical domain texts. It is more suitable for medical related tasks than traditional open world embedding\cite{chiu_how_2016}.
	The CNN model has been proved to be effective in many natural language processing tasks\cite{yin_comparative_2017}.
	
	RNN (Recurrent Neural Network) is another popular deep learning model and has achieved great success and been widely applied in many NLP \cite{cho_learning_2014} tasks.
	The purpose of RNNs is to process sequence data. 
	In the traditional neural network model, the layers are fully connected between the input layer and the hidden layer as well as the hidden layer and the output layer. 
	The nodes between each layer are disconnected.
	But this common neural network is powerless for many problems.
	For example, if you want to predict what the next word is in a sentence, you usually need to use the previous word, because the words in a sentence are not independent.
	The LSTM \cite{hochreiter_long_1997} is a special RNN model proposed to solve the problem of gradient vanishing.
	But there is a problem with modeling sentences using LSTM: it is impossible to encode information from back to front.
	Bi-LSTM\cite{schuster_bidirectional_1997} solves this problem and can better capture two-way semantic dependencies. Follow this architecture, The dynamic of a LSTM cell is controlled by an input vector ($x_t$), a forget gate ($f_t$), an input gate ($i_t$), an output gate ($o_t$), a cell state ($c_t$), and a hidden state ($h_t$), which are computed as: \\
	$$i_t = \sigma(W_i*[h_{t-1}, x_t] + b_i) \eqno(1)$$
	$$f_t = \sigma(W_{f} * [h_{t-1}, x_t] + b_f)  \eqno(2)$$
	$$o_t = \sigma(W_{o} * [h_{t-1} + b_o]) \eqno(3)$$ 
	$$g_t = \tanh{(W_g * [h_{t-1}, x_t] + b_g)} \eqno(4)$$
	$$c_t = f_t \odot c_{t-1} + i_t \odot g_t \eqno(5)$$
	$$h_t = o_t \odot \tanh{(c_t)} \eqno(6)$$
	where $c_{t_1}$ and $h_{t-1}$ are the cell state and hidden state respectively from previous time step, $\sigma$ is the sigmoid function $(\frac{1}{1+e^{-x}})$, tanh is the hyperbolic tangent function$(\frac{e^x - e^{-x}}{e^x + e^{-x}})$.
	The Bi-LSTM neural network model can automatically capture deep semantic features and classify sentences.
	
	We compare the sentence level PICO classification performance of different models. The input to the model is sentences consisting of the title and abstract of a risk model paper. Through the PICO model, the output is a tagged sentence of four labels: P, I/C, O, N, as shown in Figure \ref{fig2}. 
	
	\subsection*{Diseases Named Entity Recognition Model}
	
	\noindent The main task of the Diseases Named Entity Recognition (DNER) model is to identify and extract disease-related entities in the text. 
	Its input are the labeled sentences produced by the PICO classification model, and the output are the disease entities in the sentences.
	As we observed, most of disease entities are contained in P and O sentences. To simplify the problem, in this paper we only address the DNER task with sentences labeled with P and O. 
	The extraction task can be converted to a sequence labeling problem\cite{huang_bidirectional_2015} by assigning the annotated entities with appropriate tag representations. 
	In order to be more versatile, we use the standard ``BIO'' schema, in which each word is assigned to a label as following: B = beginning of an entity, I = inside an entity, and O = outside of an entity, and the training data are the NCBI-BIO dataset \cite{dogan_ncbi_2014}. 
	The NCBI disease corpus is fully annotated at the concept level to serve as a research resource for the biomedical natural language processing community. 
	The detailed statistical information is shown in Table \ref{table1}. \\
	\begin{table*}[h]
		\centering
		\caption{NCBI dataset statistics.}
		\begin{tabular}{|p{60pt}<{\centering}|p{60pt}<{\centering}|p{60pt}<{\centering}|p{60pt}<{\centering}|p{60pt}<{\centering}|p{60pt}<{\centering}|}
			\hline
			\textbf{} & \textbf{number of sentences}  & \textbf{avg sentence length} & \textbf{number of tokens} & \textbf{number of unique tokens} & \textbf{number of annotations} \\ \hline
			Train-NCBI & 5,576 & 23 & 132,584 & 9,805 & 2,911  \\ \hline 
			Valid-NCBI & 918 & 25 & 23,456 & 3,580 & 487 \\ \hline 
			Test-NCBI & 941 & 25 & 24,019 & 3,679 & 535 \\ \hline
		\end{tabular}
		\label{table1}
	\end{table*}
	
	In order to get better performance, our paper proposes a method to improve the model by encoding structured information of medical knowledge bases.
	Specifically, plain text contains a large amount of unstructured information, and the traditional model has low utilization rate.
	In the medical domain, there are many public structured medical knowledge bases, such as UMLS\cite{bodenreider_unified_2004}, DrugBank\cite{wishart_drugbank:_2006}, etc.
	The knowledge base can be represented into a graph. The nodes in the graph are entities, and the edges between nodes are the relationships among entities.
	For example, the semantic relationship between different medical entities is included in UMLS, which promotes the performance of the model. 
	
	We use the graph representation learning model, DeepWalk\cite{perozzi_deepwalk:_2014}, to encode the nodes in the medical knowledge base into low-dimensional space vectors.
	The DeepWalk draws on the idea of the famous word embedding model word2vec in nature language processing. 
	The basic processing element of word embedding is single word, and corresponding to the representation of network is graph node. Word embedding analyzes the implicit information of word sequence of a sentence, graph embedding captures structured information corresponding to node in the knowledge base.
	In a similar way, DeepWalk encodes the list of nodes in a random walk of a knowledge base.
	The so-called random walk can be understood in this way: It repeatedly and randomly selects the node of walking path on the network, and finally forms various of fixed-length node paths. 
	Starting from a random node first, each step of the walk is randomly selected from the edge connected to the current node. Then moving along the selected edge to the next vertex, and the process is repeated. Given a graph $G=(V, E) $, V is the node set and E is the edge set. $w_i = {v_o, v_1, \dots, v_n}$ is a sequence of several nodes, where $v \in V$.
	$v$ is a node and can not be calculated, so a mapping function $\Phi$ is introduced. The mapping function $\Phi$ maps each node in the network into a d-dimensional vector. $\Phi$ is actually a matrix with a total of $|V|\times d$ hyper-parameters which need to be updated and optimized. So the original optimization goal changed from formula 7 to formula 8. This objective function can be trained and optimized with the skip-gram model.
	$$P_r(v_n|v_0, v_1, \dots, v_{n-1}) \eqno(7) $$
	$$P_r(v_n|\Phi(v_0), \Phi(v_1), \dots, \Phi(v_{n-1})) \eqno(8) $$
	
	UMLS has nearly 3.6 million nodes and 80 million edges. In order to save training time, we filter out other languages and only keep the English version. The parameters of the training are updated appropriately on the basis of the original DeepWalk paper. The length of randomly walk list is 32, the number of nodes sequence is 10 and the word2vec window size is 5. The training takes 2 days on a computer with 128G RAM and 32 core CPU. We use the downstream task, NER, to evaluate the quality of the entity embedding, which in turn tunes the training of the DeepWalk.
	After that we get the medical knowledge base entity embedding and we use the knowledge graph information the same as the way of word embedding.
	For comparison, we implement different models, such as Bi-LSTM, Bi-LSTM + CRF\cite{huang_bidirectional_2015} and conduct full experiments to prove the effectiveness of the methods. For the credibility of the experimental results, the hyper-parameters are consistent with relevant papers.
	
	\subsection*{Mapping Model}
	
	\noindent The mapping model evaluates and adjusts the negative results from the previous steps. 
	Intuitively the following issues are highlighted: 
	First, DNER based on P-labeled or O-labeled sentences would get empty result. 
	The solution of this issue is to look for the results in the I/C-labeled and N-labeled sentences. 
	Second, there is an intersection between the recognition results of P and O which we should keep only one if an entity appear in both of P and O at the same time. 
	The solution is to leverage the ``soft'' label of the PICO sentences classification model which are the probabilities of corresponding classes on each sentence. As we mentioned previously, a sentence might contain more than two PICO elements. For example, the first half of sentence belongs to P label while the latter part is belonging to O label. The PICO classification model cannot solve this problem.
	In order to solve this issue, we incorporate the method of linguistic rules. The final score calculation formula for disease entity is as follows:
	$$ Score = \lambda f(d_i) + (1-\lambda)g(d_i) \eqno(9)$$
	where $d_i$ represents a disease entity which has two scores $d_i = (s_1, s_2)$. 
	$s_1$ represents the probability that the entity belongs to P label, and $s_2$ represents the probability that the entity belongs to O label. 
	
	This probability is generated from the aforementioned PICO classification model.
	$g(d_i)$ stands for the regular expression rule model. We manually formulate some typical linguistic rules, which are deterministic for the segmentation of sentence, e.g., if they belong to P or O. If the disease entity $d_i$ is covered by the rule, then the value of $g(d_i)$ is set to 1, otherwise 0. The results of g(s) are (1,0), (0,1) and (0,0), respectively.
	Re-dividing the entity associated label set based on the score can improve the overall performance and reduce the effects caused by the error accumulation of the previous models. In order to verify the method, we invite several medical experts to help us label 100 articles. We compare the results of adding rules with no rules and calculate the recall rate.
	
	\section*{Results}
	\indent In order to fully demonstrate the effectiveness of our method, we conducted complete experimental evaluation for each module, and obtained convincing experimental results. Precision, Recall and F1 score are used as indicators which are the most commonly used measurements in the field of machine learning and deep learning.
	
	In the area of information retrieval, true positive (TP) indicates the number of labeled instances the method find. False positive (FP) indicates the number of unlabeled instances that the method find. And false negative (FN) indicates the number of labeled instances that the method doesn't find. Then precision, recall and F1 score are calculated as following: 
	$$ Precision = \frac{TP}{TP+FP} \eqno(10) $$ 
	$$ Recall= \frac{TP}{TP+FN} \eqno(11) $$
	$$ F1 = \frac{2*Precision*Recall}{Precision+Recall} \eqno(12)$$
	
	\subsection*{Results of PICO Classification Model}
	
	\noindent The 500 papers for the PICO classification model are downloaded from PubMed sorted by time. Two experts are invited to label the same batch of papers at the same time independently. If there is a disagreement, it will be confirmed by the third expert. We finally get our dataset and divide it into train set and validation set with a 7:3 ratio. In order to ensure the adequacy of the experiment, two classification models are implemented: CNN \cite{kim_convolutional_2014} and Bi-LSTM \cite{graves_framewise_2005}. However, the scores of the Bi-LSTM model are much higher than the CNN model's, so here we only show the result of the Bi-LSTM model as in Table \ref{table2}. We compare our classification results with Jin et al.\cite{jin_pico_2018}, which extracted 489k medical abstracts from MEDLINE(https://www.medline.com/). Yuan et al. \cite{yuan_extracting_2019} proposed a soft-margin SVM classification model which merges other features that combines 1-2 gram analysis with TF-IDF\cite{pan_survey_2010} method. We also compare with them.
	
	\begin{table*}[h]
		\centering
		\caption{Comparison of PICO classification models.}
		\begin{tabular}{|c|c|c|c|c|c|c|c|c|c|}
			\hline
			\multirow{2}{*}{} & \multicolumn{3}{c|}{P}                           & \multicolumn{3}{c|}{I/C}       & \multicolumn{3}{c|}{O}         \\ \cline{2-10} 
			& Precision              & Recall              & F1             & Precision     & Recall             & F1    & Precision     & Recall              & F1    \\ \hline
			Jin2018\cite{jin_pico_2018}          & 0.885          & 0.828          & 0.856          & 0.749 & 0.815          & 0.781 & 0.845 & 0.832          & 0.838 \\ \hline
			Yuan2019 \cite{yuan_extracting_2019}         & 0.925          & 0.838          & 0.879          & 0.842 & 0.789          & 0.814 & 0.886 & 0.897          & 0.891 \\ \hline
			Our Model (Bi-LSTM)         & \textbf{0.947} & \textbf{0.867} & \textbf{0.906} & 0.773 & \textbf{0.850} & 0.810 & 0.867 & \textbf{0.918} & \textbf{0.891} \\ \hline
		\end{tabular}
		\label{table2}
	\end{table*}
	
	\subsection*{Results of DNER Model}
	\noindent The performance of the DNER model has a major impact on the final results. 
	We compare the performance of different models and the same model under different hyper-parameters. Most of the hyper-parameters follow Reimers  \cite{reimers_reporting_2017}: dropout=0.25, lstm hidden layer size = 100, optimizer=adam, char cnn embedding size=30, char LSTM embedding size=25, and mini batch size=32.
	In addition, in order to improve performance, we introduce structured information in the medical knowledge base via graph representation model training and prove its effectiveness from experimental results as shown in Table \ref{table3}.
	We also use the BERT model which is the dominated model in various fields for medical entity recognition tasks.
	The experimental code refers to BERT-pytorch (https://github.com/codertimo/BERT-pytorch). 
	Two different versions of BERT are implemented: BERT-base and BERT-large. With fewer parameters, BERT-base has faster training speed. but BERT-large achieves the best performance over all the other models with precision 0.8507, recall 0.8844 and F1 score 0.8672.
	\subsection*{Results of Mapping Model}
	\noindent The mapping model is used to perform further processing of the DNER model and obtain the disease entity classification results, so that the clinical practitioners can quickly get the key evidence with less time. Several clinical researchers read series of medical papers of risk models and summarize a set of linguistic rules such as ``... risk of $<outcome>$''. If a sentence matches this rule, the entity that appears in the later part is extracted and classified as ``O''. 100 papers are labeled by the clinical researchers following the same protocol with the PICO classification labeling work. The results are shown in Table \ref{table4}.
	\begin{table*}[h]
		\centering
		\caption{Result of different DNER models.}
		\begin{tabular}{|c|c|c|c|}
			\hline
			model                              & Precision               & Recall               & F1              \\ \hline
			LSTM+SOFTMAX                       & 0.7910          & 0.8125          & 0.8016          \\ \hline
			LSTM+CRF                           & 0.8179          & 0.8375          & 0.8275          \\ \hline
			LSTM+CRF+CNN CHAR                  & 0.8298          & 0.8333          & 0.8316          \\ \hline
			LSTM+CRF+LSTM CHAR                 & 0.8283          & 0.8697          & 0.8485          \\ \hline
			LSTM+CRF+CNN CHAR+GRAPH EMBEDDING  & \textbf{0.8433} & 0.8583          & \textbf{0.8508} \\ \hline
			LSTM+CRF+LSTM CHAR+GRAPH EMBEDDING & 0.8348          & \textbf{0.8635} & 0.8489          \\ \hline
			BERT BASE                          & 0.8200          & 0.8729          & 0.8456          \\ \hline
			BERT LARGE                         & \textbf{0.8507} & \textbf{0.8844} & \textbf{0.8672} \\ \hline
			Zhai, Zenan 2018                   & 0.8213          & 0.8366          & 0.8289          \\ \hline
			TaggerOne (Leaman and Lu, 2016)    & 0.852           & 0.802           & 0.826           \\ \hline
			Dnorm (Leaman et al., 2013)        & 0.820           & 0.795           & 0.807           \\ \hline
		\end{tabular}
		\label{table3}
	\end{table*}
	
	\begin{table*}[h]
		\centering
		\caption{Result of Mapping Model with different method.}
		\begin{tabular}{|c|c|c|}
			\hline
			\multirow{2}{*}{Model}               & \multicolumn{2}{c|}{Recall}      \\ \cline{2-3} 
			& P              & O               \\ \hline
			Bi-LSTM+CRF+CNN CHAR+GRAPH EMBEDDING & 0.7262         & 0.60            \\ \hline
			BERT LARGE                           & 0.7588         & 0.6293          \\ \hline
			BERT LARGE+ LINGUISTIC RULE            & \textbf{0.7911} & \textbf{0.7474} \\ \hline
		\end{tabular}
		\label{table4}
	\end{table*}
	\subsection*{Web Application}
\begin{figure*}[h!]
		\centering
		\includegraphics[scale=1, width=430pt, height=300pt]{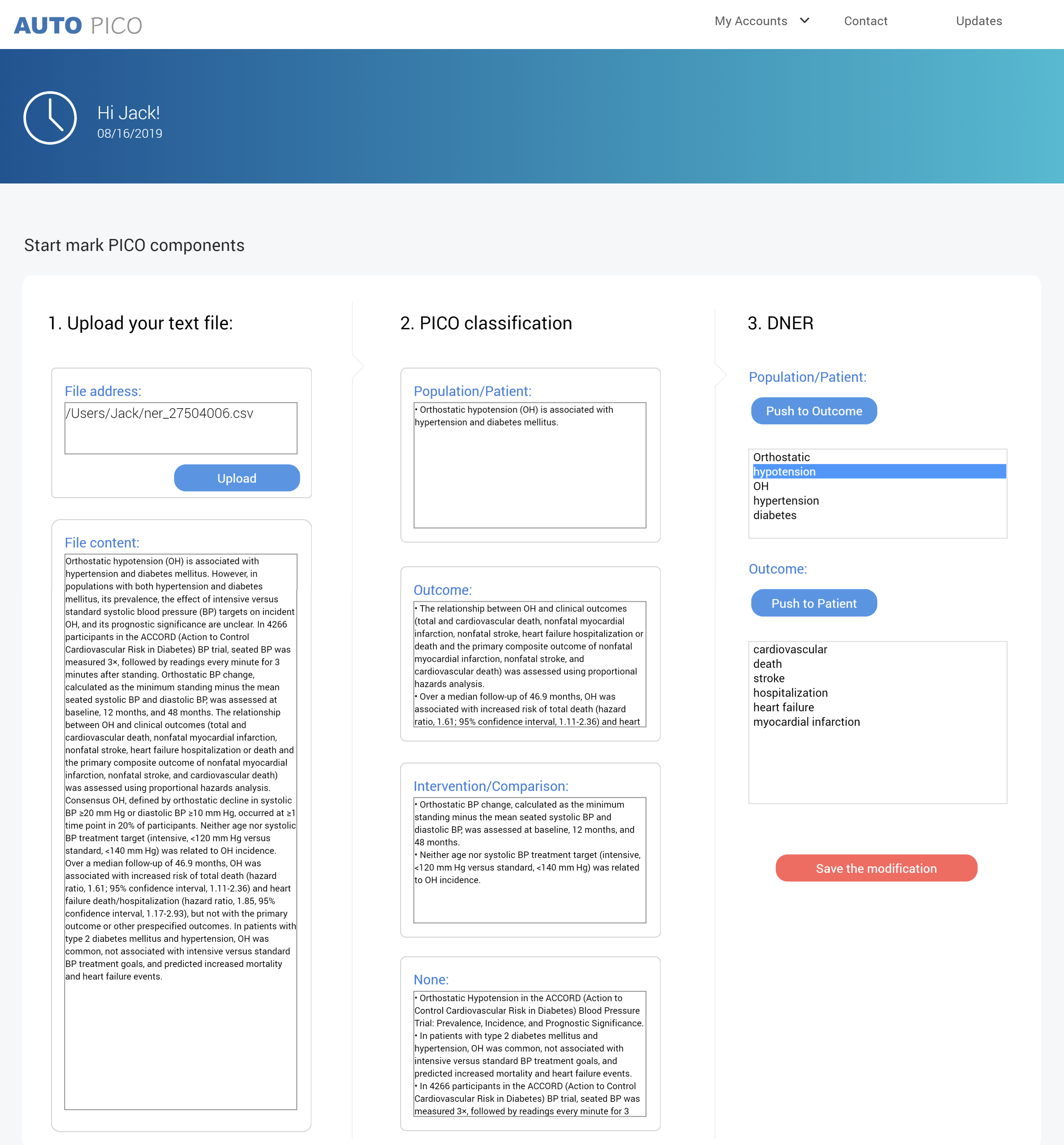}
		\caption{The visualization System.}
		\label{fig3}
	\end{figure*}

	\indent A web application is developed to enable the method and provide interface (see Figure \ref{fig3}) for user to interact with the results. When the user uploads a paper, the back-end system will call the service API on the server to parse the text in title and abstract, and display the PICO classification result and the corresponding disease entity recognition result. At the same time, to further improve the accuracy of the system, we also provide labeling and modification buttons. If there is an error in the recognition result, the user can manually modify the result and store it in the back-end database. If the system collects very few samples, it will use the language rules and remember it to avoid mistakes next time. When samples are accumulated to a certain extent, the back-end system will retrain the model to improve the performance of the model. 
	\noindent As in Figure \ref{fig3}, the title and abstract of the risk model paper are displayed on the left side of the web page, the panel in the middle shows the results of the sentence-level PICO classification. The right panel shows the final disease entity mapping results in P and O respectively. 
	\section*{Discussion}
	\noindent The methods mentioned earlier provide a step-wise framework for automatically extracting disease entities from risk model papers based on the PICO framework. The following sections analyze and discuss some details of the experimental results for each module.
	
	In table \ref{table2}, we compare the PICO classification results of different models based on Bi-LSTM architecture. Our model achieves the best performance on the P label sentences classification comparing with the state-of-the-art models as well as some indicators of I/C and O labels, which proves the effectiveness of our model.
	In addition to the performance comparison, we also analyze the negative samples. Statistics show that in some cases, due to the frequent occurrence of some typical words, the sentence of P label is difficult to distinguish from I/C, the same as O and I/C. In addition, in some sentences, the part of one sentence belongs to P label, however, the other part belongs to O label. This increases the difficulty of classification. We compare with the results of two recent related papers, and we can clearly see that our results in most indicators are better.
	
	In the DNER module, we experiment with leading deep models on NER task (Table \ref{table3}). For models with Bi-LSTM, our model Bi-LSTM + CRF + CNN char + graph embedding improves the baseline of F1 score from $80.16\%$ to $85.08\%$ and precision score from $79.10\%$ to $84.33\%$. 
	Compared to current state-of-the-art models, we achieve significant improvements.
	
	The BERT model outperforms all other models on all the measurements (the precision score is 85.07\%, the recall score is 88.44\%, the F1 score is 86.72\%). What we can not ignore is that the model is too complex and the calculate process is very slow, so each time the parameter adjustment takes a lot of time.
	
	For the mapping model, since some sentences may have key elements of P and O at the same time, classification is difficult and has a great impact on the results of DNER. The mapping model uses a set of statistical rules and language rules to alleviate this problem and improve the overall performance.
	In table \ref{table4}, we only give results about recall because we care more about how many entities are found which we need. The precision actually is low due to following issues: (1) there are still errors from the PICO classification model and the DNER model which have not been well addressed, and (2) there's lack of public standard labeling strategies. 
	Note that the manual work of labeling disease entities with PICO elements is a non-trivial work for most of the clinical researchers even we restrict the papers with risk model type. During the experiment, we observe that a good deal of labels done by the first two researchers are mismatched. That also increase the work of the third researchers to further do a decision. By considering this situation and less number of labeled samples, we finally decide to use rule-based method in the mapping model. Specifically for risk model disease entity extraction and classification, this is the first try to do such experiments and can serve as a benchmark for future research. 
	
	One advantage of step-wise approach is that the results of each step can be output and presented to the user. In Figure \ref{fig3}, we show an example with the medical paper \cite{fleg2016orthostatic} which is randomly retrieved from PubMed. The middle of the web page is the classification result of PICO elements and the DNER recognition results are shown on the right panel. It can be found that the classification result of PICO sentences is reasonable, but the results of DNER have some errors. For example, the item ``hospitalization'' should not belong to the disease entity caused by the misclassification of the DNER model. At this point, the user can continuously improve the performance of the model by deleting this unreasonable word and saving the result to the back-end system.
	
	\section*{Conclusion}
	In this paper, we propose a step-wise method for extracting medical entities based on the PICO framework. Main steps include PICO sentence classification, disease entity recognition and disease mapping. Mainstream deep neural networks such as CNN and Bi-LSTM are used to classify sentences into PICO elements. The disease entity recognition is based on dominated deep learning frameworks. Structural medical knowledge is embedded with probabilistic knowledge mapping model. Experimental results show that our method achieves reasonable performance on precision, recall and F1 score. An online web system is also developed to facilitate clinical researchers to do medical entity based PICO extraction.
	
	\setlength\itemsep{-0.1em}
	\bibliographystyle{aaai}
	\bibliography{AAAI_PICO_paper_2019}
\end{document}